\batchmode
\makeatletter
\def\input@path{{C:/Users/tanj0/Desktop/MASpaper-master-journal/journal/}}
\makeatother
\documentclass[english,aic]{iosart2x}
\usepackage[utf8]{inputenc}
\usepackage{array}
\usepackage{multirow}
\usepackage{graphicx}

\makeatletter

\DeclareTextSymbolDefault{\textquotedbl}{T1}
\providecommand{\tabularnewline}{\\}

\usepackage{graphicx}        
\usepackage{multicol}        

\usepackage{listings}
\usepackage{color}
\usepackage{hyperref}

\definecolor{dkgreen}{rgb}{0,0.6,0}
\definecolor{gray}{rgb}{0.5,0.5,0.5}
\definecolor{mauve}{rgb}{0.58,0,0.82}

\pubyear{2020}
\volume{0}
\firstpage{1}
\lastpage{1}

\makeatother

\usepackage{babel}
\begin{document}
\makeatletter
\let\put@numberlines@box\relax
\makeatother
\begin{frontmatter}
\title{A Novel Multi-Agent Scheduling Mechanism for Adaptation of Production
Plans in Case of Supply Chain Disruptions}

\maketitle
\runningtitle{A Novel Multi-Agent Scheduling Mechanism for Adaptation of Production Plans in Case of Supply Chain Disruptions}

\author[A]{\inits{J.}\fnms{Jing} \snm{Tan}\ead[label=e1]{jingtan@huawei.com}
\thanks{Corresponding author. \printead{e1}.}}
\runningauthor{J. Tan et al.}
\author[B]{\inits{L.}\fnms{Lars} \snm{Braubach}\ead[label=e2]{lars.braubach@hs-bremen.de}
\thanks{Corresponding author. \printead{e2}.}}
\author[C]{\inits{K.}\fnms{Kai} \snm{Jander}\ead[label=e3]{jander@th-brandenburg.de}
\thanks{Corresponding author. \printead{e3}.}}
\author[D]{\inits{R.}\fnms{Rongjun} \snm{Xu}\ead[label=e4]{xurongjun@huawei.com}}
\author[E]{\inits{K.}\fnms{Kai} \snm{Chen}\ead[label=e5]{colin.chenkai@huawei.com}}
\address[A]{\orgname{Huawei Technologies Co. Ltd.},
\cny{China}\printead[presep={\\}]{e1}}
\address[B]{\orgname{City University of Applied Sciences Bremen and Actoron GmbH},
\cny{Germany}\printead[presep={\\}]{e2}}
\address[C]{\orgname{Brandenburg University of Applied Sciences and Actoron GmbH},
\cny{Germany}\printead[presep={\\}]{e3}}
\address[D]{\orgname{Huawei Technologies Co. Ltd.},
\cny{China}\printead[presep={\\}]{e4}}
\address[E]{\orgname{Huawei Technologies Co. Ltd.},
\cny{China}\printead[presep={\\}]{e5}}

\begin{abstract}

Manufacturing companies typically use sophisticated production planning
systems optimizing production steps, often delivering near-optimal
solutions. As a downside for delivering a near-optimal schedule, planning
systems have high computational demands resulting in hours of computation.
Under normal circumstances this is not issue if there is enough buffer
time before implementation of the schedule (e.g. at night for the
next day). However, in case of unexpected disruptions such as delayed
part deliveries or defectively manufactured goods, the planned schedule
may become invalid and swift replanning becomes necessary. Such immediate
replanning is unsuited for existing optimal planners due to the computational
requirements. This paper proposes a novel solution that can effectively
and efficiently perform replanning in case of different types of disruptions
using an existing plan. The approach is based on the idea to adhere
to the existing schedule as much as possible, adapting it based on
limited local changes. For that purpose an agent-based scheduling
mechanism has been devised, in which agents represent materials and
production sites and use local optimization techniques and negotiations
to generate an adapted (sufficient, but non-optimal) schedule. The
approach has been evaluated using real production data from Huawei,
showing that efficient schedules are produced in short time. The system
has been implemented as proof of concept and is currently reimplemented
and transferred to a production system based on the Jadex agent platform.\end{abstract} 

\end{frontmatter}

\section{Introduction}

Companies who have experienced severe supply chain disruptions often
report substantial revenue losses and reduced market value after
the event \cite{craighead2007severity}\cite{tomlin2006value}. Hendricks
et al. \cite{hendricks2005empirical} show that such disruptions are
likely to have long-term financial impact.

Mitigation and contingency tactics range from passive acceptance to
active customer demand steering \cite{tomlin2006value}. A company's
decision is often an attempt at balancing multiple objectives, when
critical decision factors are unclear and causal effects are hard
to explain. Vakharia et al. \cite{vakharia2009managing} introduces
various MIP (Mixed-Integer Programming) models, analytically solving
for important decision points such as lead time and safety stock \cite{vidal1997strategic},
preventive supply chain partner selection \cite{gaonkar2003robust},
optimized material flow (order volume and source) with reliability/robustness
constraints \cite{bundschuh2003modeling} and a global supply chain
network design \cite{kouvelis2008structure}. Due to growing complexity
in a global supply network, traditional optimization models rely on
expert knowledge regarding critical factors in the model as well as
the fine-tuning required to overcome potentially huge result variations.

\begin{center}
\begin{figure*}
\begin{centering}
\includegraphics[width=1\textwidth]{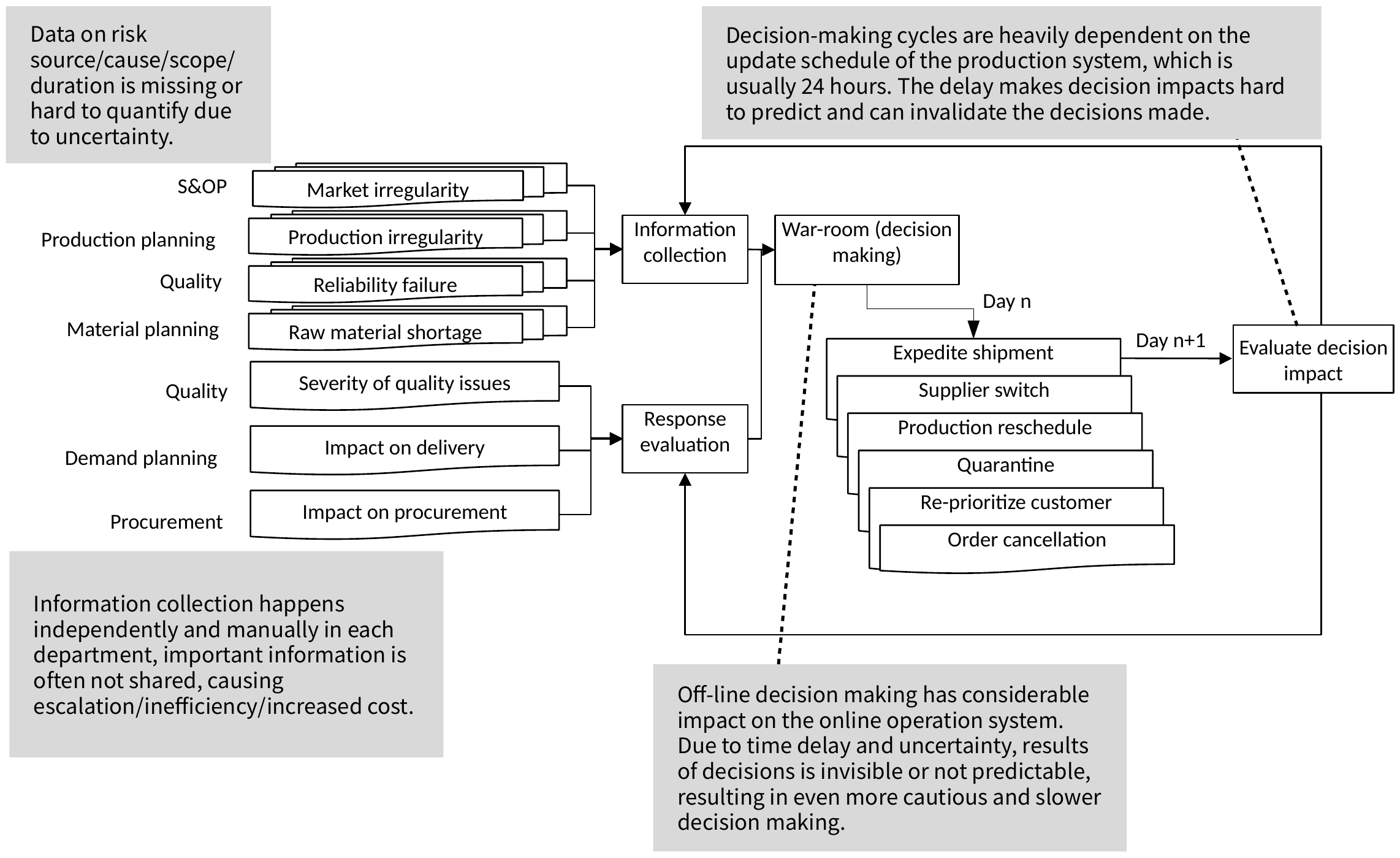}
\par\end{centering}
\caption{Current process at Huawei to compensate for supply chain disruptions\label{fig:oldscenario}}
\end{figure*}
\par\end{center}

This study is based on a project of the supply chain department of
Huawei Technologies Co. Ltd, an international manufacturing company.
Huawei manufactures a large number of complex products which in turn
consist of many subcomponents. The subcomponents themselves can either
be purchased from a supplier but are also themselves manufactured
by Huawei out of even simpler parts. Each of the subcomponents require
different inputs and manufacturing processes, resulting in a complex
supply chain for each of the final products being manufactured. The
supply chain includes multiple suppliers, production facilities and
associated logistics to ensure that components are available when
they are needed and orders can be fulfilled. However, this complexity
also results in the supply chain being vulnerable to disruptions:
For example, suppliers may suddenly be unable to deliver components,
logistical issues may delay or prevent components from arriving, unexpected
breakdowns at manufacturing facilities result in delays or require
other facilities to substitute for the failed one.

In addition, some problems may not be immediately apparent: Components
may not reach the mandated level of quality or be produced with flaws
that are not immediately discovered. When the flaws become apparent,
a production batch that has already been produced may have to be quarantined
to prevent the flaw from causing quality issues in products down the
supply chain. This quarantine leads to an immediate supply shortage
that needs to be compensated.

Furthermore, all production batches of both final and intermediate
products have production deadlines, the deadlines of intermediate
products being driven by the ones of the final products. Simply delaying
the delivery of a production batch may result in contractual fines
or outright rejection of the production batch by the customer, resulting
in financial losses. These deadlines create an urgency in case of
suddent disruption of the supply chain that require immediate action
to avoid those financial losses.

Since the normal production planning cycle is 24 hours, which may
take too long, manual processes are often used to develop steps to
compensate the disruption such scenarios. Figure \ref{fig:oldscenario}
shows the current manual approach used by Huawei in case of unexpected
supply chain disruptions: In case of a disruption, a so-called ``war
room'' of experts is created with the aim of finding an immediate
solution adjusting the production plans to compensate for the supply
shortage caused by the disruption: First, the war room collects all
available information. This process step is already impeded by manual
information collection and outright missing information about current
production runs and schedules. This information often has to be collected
manually right from the production facility shop floor.

Once acquired, the war room decision makers use the information to
propose changes to the production schedule, for example quickly setting
up an extra production run at a different facility. However, such
schedule changes in complex supply chains like this are likely to
cause ripple effects: The newly-tasked facility is now unavaible for
other production runs and may have to delay runs that have already
been scheduled. As a result, when the changes are then fed into the
standard production planning system, which then takes the changes
into account within the next planning cycle, it will show the impact
of the changes implemented by the war room on the global supply chain.
The war room can then use this information to yet again implement
further changes, creating a feedback loop.

Due to ripple effects, many war room actions may be required, however,
due to the large feedback delay of 24 hours, the process very cumbersome
and difficult, with the impact of changes only visible after a substantial
delay. This leads to the results of war room decisions being hard
to predict and the decision makers often becoming overly cautious
when deciding on necessary scheduling changes. This means that they
are often unable to implement bold changes to ``save'' a production
order or are only gradually closing in on a solution over a long time
using the feedback loop.

The company was subject to several supply chain disruptions in the
past year. \textquotedbl War rooms\textquotedbl{} were used during
such disruptions to share information across functional departments,
agree on decisions based on the status quo, wait until the next day
for decision effects to materialize and incorporate new inputs to
devise further decisions. This iterative process is manual, inefficient
and cumbersome to the point of delaying critical decisions.

The goal of this study is not to design a system replacing existing
planning and scheduling systems, but instead offers a mechanism, which
closely resembles the behavior of such systems, imitating their response
at a fraction of the original time at the expense of a slightly reduced
solution quality. Typically the standard scheduling system requires
hours to reach an optimal solution. If used for simulating various
scenarios, it may necessitate days to collect and compare results.
This study prototypes a simulation system which enables 1) playing
through scenarios in minutes, eliciting responses similar to the actual
system supporting urgent management decisions in emergency situations
and 2) simulating a diverse set of disruptions of varying severity
in advance supporting preventive supply chain design changes.

The study proposes a hybrid approach, solving multiple simple MIP
problems in parallel using decentralized multi-agent system. Each
agent controls its own resources and is responsible for its own individual
and independent objectives. In addition, its visibility is limited
to its immediate suppliers and customers (neighboring agents). During
disruptions, directly affected agents will start optimizing their
individual objectives with simple MIP techniques based on reduced
resources (capacity or material); resulting interim results will propagate
along the chain to their suppliers or customers in stages aiming to
minimize number of nodes which need to deviate from the initial state.
In case of conflicts, i.e. a proposal cannot meet the agent's internal
constraints, the agent will optimize and propose an alternative solution,
then propagate it back through the chain. The system achieves stability
when all agents meet their constraints. Since MIP optimization independently
occurs inside each agent, the number of parameters and constraints
are substantially reduced compared to modeling and optimizing the
entire system. Despite increased communication complexity, the distributed
system rapidly stabilizes.

This paper is an extended version of an IDC symposium paper \cite{TanXuChenBraubachJanderPokahr2019}
and especially adds background information about the Huawei use case
scenario and explains the system realization. It is structured as
follows: Section 2 provides a brief overview of current simulation
methods. Section 3 describes the system prototype developed in the
project. In Section 4 the implementation is outlined. Thereafter,
in Section 5 simulation results with sample data are provided. Finally,
in Section 5, conclusions are drawn and further research areas are
suggested.

\section{Related Work}

This section presents a broad overview of simulation techniques and
agent-based approaches in the context of scheduling problems.

\subsection{Simulation Techniques}

A common dynamic modeling technique is systems dynamics modeling (SD).
As described by Li et al. \cite{li2016system}, the modeling approach
adjusted by supply chain risk modeling highlights causal inter-dependencies
and interactions. Uncertainty is captured through probability distribution
of parameters; causal relation (including time delays) is expressed
by mathematical equations. Using sampled parameter values and running
a Monte-Carlo simulation, worst-, average- and best-case scenarios
can be derived. Although suitable for capturing interactions, SD depends
on expert knowledge of system structure and parameters \cite{weimer2006cross}\cite{ghadge2013systems}.

Discrete event simulation (DES) is another prevailing modeling technique
in logistics and supply chain management. In contrast to SD, it models
state changes in discrete time steps and entities are represented
individually. In literature review by Tako et al. \cite{tako2012application},
both simulation techniques are compared. In conclusion, despite both
approaches being used extensively as decision support tools, DES is
more suitable for operational/tactical level modeling such as local
production decisions, whereas SD is more appropriate for long-term
and strategic modeling. Choice of modeling technique relies on balancing
data volume and complexity of the system structure: in general, a
DES system structure requires less expert knowledge while SD requires
less data.

\subsection{Agent-Based Modeling}

Compared to SD or DES, agent-based simulation relies on bottom up
modeling of supply chain roles by heterogeneous agents acting autonomously,
often using simple rules but expressing behaviors as a system not
explicitly programmed (emergent behavior) \cite{macal2005tutorial}.
A common way of designing agents uses different roles and functionalities
in the supply chain. For example, Ledwoch et al. \cite{ledwoch2018moderating}
model supply chains with supplier agents, OEM agents and logistic
provider agents; Otto et al. \cite{otto2017modeling} employ a similar
approach to model response dynamics of a system under production shocks.
Such models generally focus on studying complex supply network topology,
material and information flow but are deficient modeling operation-level
planning and scheduling. They simplify product hierarchy with one
or a few dummy products, and rely on general assumptions like time
delay to imitate planning processes and capacity limits within each
agent. Seck et al. \cite{seck2015simulation} apply a more operative
approach to separate agents into product, demand, production, stock,
order and batch entities. This approach allows studying networks with
more operational aspects such as Bill-Of-Materials (BOM), forecast
and firm demands, capacity and optimal production batches, etc. Although
tested with a limited number of nodes and products, its model is extensible
for more intricate product hierarchies.

Current level of modeling detail in existing approaches is insufficient
for daily operational planning and scheduling. These approaches are
inadequate for short-term supply chain disruption simulations of individual
material codes and production capacity, especially when decisions
involve more than 10,000 material codes and hundreds of production
lines in various manufacturing locations. Our approach uses agent-based
modeling and simulation techniques in which agents are modeled on
an even lower level than Seck et al. \cite{seck2015simulation}. Here,
each agent represents one specific type of product or production capability.
Scheduling of competing products for limited material availability
and production capability is done using iterative communication and
compromises between agents.

\begin{figure*}
\begin{centering}
\includegraphics[width=1\textwidth]{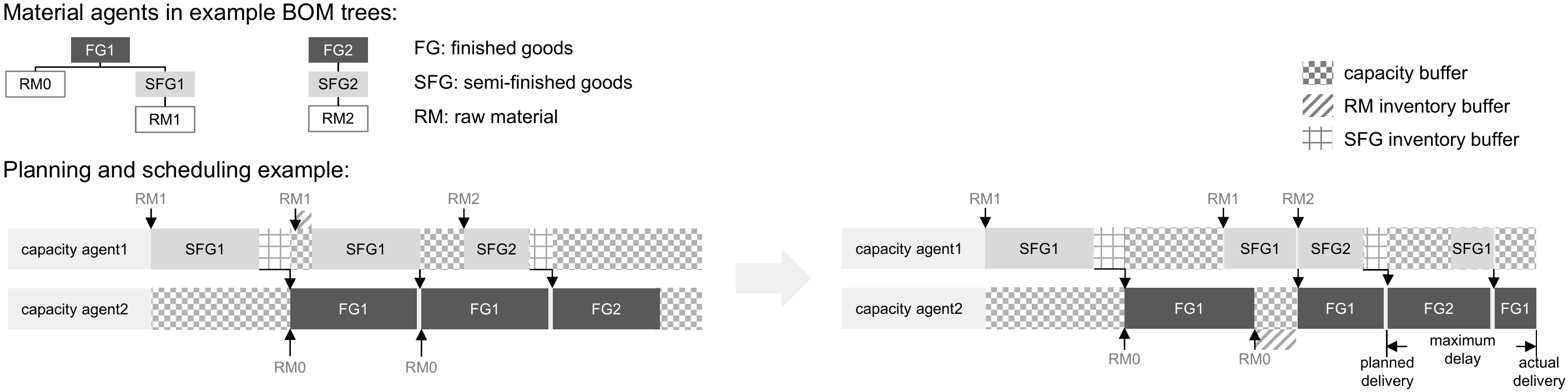}
\par\end{centering}
\caption{An illustration of rescheduling \label{fig:samplematagents}}
\end{figure*}
Since this study is focused on disruptions, communication and compromise
always leads to reduced or delayed production compared to the baseline.

The underlying idea of the approach proposed in this paper consists
in having agents that represent important parts of the system with
their individual world view and goals. These agents use negotiations
to achieve their individual goals and efficiently generate schedules
that modify an existing solution. Due to the more limited local data
sets, this kind of scheduling in unable to achieve better solutions
compared to the optimized quality of results of a global planner.
Nevertheless, it is possible to achieve other quality criteria such
as high performance (due to more limited local problem spaces and
parallelism), any-time solutions (using iterative improvement), few
changes to an existing schedule etc. In this respect the approach
proposed in this paper shares some similarities with an agent-based
hospital logistics simulation approach and system originally developed
by some of the authors as part of previous works called MedPAge (medical
path agents) \cite{Paulussen+06Agent-based}. This approach is similar
to the proposed system in that it contains agents providing resources
(i.e. hospital services) to agents requiring those resources (patients).
It also shares a negotiation-based approach towards finding a solution:
In this scenario patients and hospital resources like an X-ray machine
are represented as agents, while the first group intents to improve
their health state as fast as possible and reduce stay time, the latter
group prefers to have high utilization of hospital resources. Both
types of agents negotiate with each other for appointment and treatment
times and can further optimize an initial schedule into a pareto-optimal
solution by exchanging received slots. However, the motivations and
goals for the system were different from the current proposal in that
organizational structures and practices played a significant role
while the current approach is aimed at assisting in swiftly handling
sudden production disruptions.

\section{Agent-based Disruption Scheduling Approach}

\begin{figure*}
\begin{centering}
\includegraphics[width=1\textwidth]{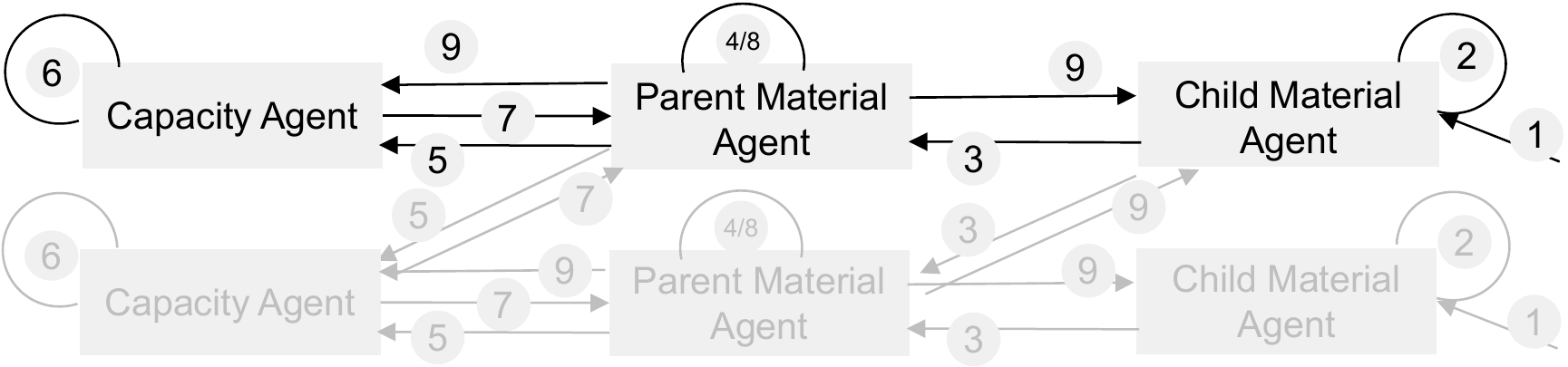}
\par\end{centering}
\caption{Interaction between agents\label{fig:interaction}}
\end{figure*}

The approach is based on real-world Bill-Of-Materials (BOM). A BOM
is a tree or network structure representing relationships between
raw materials, semi-finished and finished goods. Leaf nodes on the
bottom of the tree or network are the lowest-level raw materials and
root nodes at the top are finished goods for customers. Material codes
on the lower layer need to be produced or purchased before their parent
(downstream customer) nodes can be produced. Scheduling complexity
can be defined by the number of layers and connections BOMs: BOMs
with copious layers and connections imply high levels of inter-dependence
between material codes and therefore result in more complex scheduling.

Two types of agents used, representing entities with different capabilities
in the supply network. A \emph{material agent's} objective is producing
a designated type of material in a specific location to satisfy as
many of its downstream customer agents' orders as possible, on time
and in full. It has the resources of in-stock and in-transit supply
inventory. It controls its own production schedule and has partial
visibility into its immediate upstream and downstream agents' schedules
when announcing new proposals concerning its own material. It also
includes information of substitute materials for its standard-choice
of supply. It can communicate with other agents through a protocol
for sending and receiving new scheduling proposals; it can optimize
its production schedule or alternatively delay, cancel or combine
production orders. It is initialized with a predefined production
schedule.

The \emph{capacity agent} represents the machinery used to produce
goods. Like a real-life production line, capacity agents deal with
so-called \textquotedbl capacity packages\textquotedbl , representing
a group of semi-finished or finished goods with similar characteristics
so that they can be produced using the same production line in only
slightly different configurations. These agents' objective is to fulfill
the maximum number of capacity orders on time and in full. A major
difference between a material agent and a capacity agent is that unused
capacity left at the end of each planning time unit will be deleted.
In case of a resource shortage it can optimize its production and
propose new schedules.
\begin{center}
\par\end{center}

Fig. \ref{fig:samplematagents} (upper part) demonstrates two simple
BOM trees modeled using 7 material agents, 4 of which impose requirements
of production capacity on 2 capacity packages (represented by 2 capacity
agents). The initial production schedule, as shown in Fig. \ref{fig:samplematagents}
(lower left part) has built-in inventory and capacity buffers to cope
with short-notice schedule changes. Disruption lasting no longer than
that buffer length require no rescheduling. If it occurs on short
notice lasting for a longer period of time, the planner will attempt
rescheduling so that the existing production sequence and amount necessitate
minimum change. As an example shown in Fig. \ref{fig:samplematagents}
(lower left part), the second delivery of RM1 is delayed long enough
to affect scheduling of other material codes. Fig. \ref{fig:samplematagents}
(lower right part) shows one potential rescheduling solution where
production of affected SFG1 and FG1 is split.

\subsection{Conceptual Overview}

Fig. \ref{fig:interaction} illustrates the interaction between agents
based on a single rescheduling iteration. When a disruption occurs,
system input data changes to reflect reduced amount of raw materials
and/or semi-finished goods. As a first step (cf. \ref{fig:interaction}),
this only triggers immediately affected agents. In response, in-stock
and in-transit inventory is recalculated. In step 2, triggered agents
attempt to reschedule and optimize their production schedules using
the new information. The optimization problem is modeled as a simple
MIP problem. Due to the agent's limited visibility and control scope,
the potentially complex system-wide optimization problem is reduced
to individual agents resolving their own small problems. After solving
their own optimization problems, agent propose changes to their downstream
customer agents' schedules. In step 4, the affected customer agents
consolidate all change proposals received from their upstream suppliers
and optimize their own schedule.

In general, material agent schedules can only degrade under the assumption
that supply chain disruption risks always negatively impact the plan.
Proposed changes will be propagated to related capacity agents. In
steps 5 to 7, communication and rescheduling will be initialized by
affected capacity agents. A capacity agent receives change proposals
from all relevant material agents and attempts to solve its own optimization
problem. Steps 8 and 9 again describe consolidation of new proposals
and the corresponding rescheduling activities. The current simulation
system only models disruption scenarios. However, it is straightforward
to extend to allow simulating human interference such as order priority
changes, capacity increases or expedited shipments.

\subsection{MIP Problem}

Step 2 of the rescheduling iteration includes an optimization problem
within the supply material agent (i.e. a child node on a lower level
in the BOM) (cf. Section \ref{subsec:Supplier-Material-Agent's}).
There are two options used in the simulation: if partial order fulfillment
is permitted, agents attempt to produce as early and as much as possible,
even if it means full amounts will be produced separately or partially
cancelled. If partial order fulfillment is prohibited, agents try
to produce orders either in full or not at all. This behavior is modeled
after real-world scenarios where customer demands need to be produced
and shipped as single batch or completely cancelled by the customer.
In step 4, a customer material agent (i.e. a parent node on a higher
level in the BOM) solves the optimization problem (cf. Section \ref{subsec:Customer-Material-Agent's}).
It is common source of multiple suppliers who independently delay
their material delivery. The plant reschedules its production accommodating
all potential delays in one shot instead of dealing with each disruption
separately. In step 6, a capacity agent solves its optimization problem
(cf. Section \ref{subsec:Capacity-Agent's-Consolidation}). After
the system stabilizes, an optional inventory reduction step strips
off excessive inventory caused by order cancellation induced by local
optimizations in all material agents. This step is not described in
detail in this paper.

\subsection{Material Agent \label{subsec:Material-Agent}}
\begin{center}
\begin{figure*}
\begin{centering}
\includegraphics[width=1\textwidth]{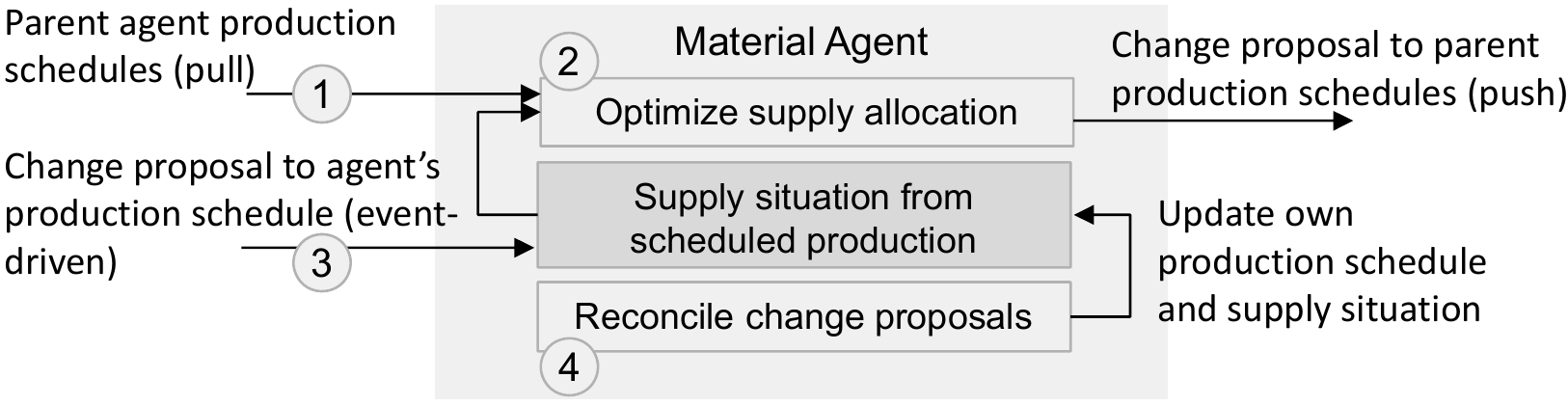}
\par\end{centering}
\caption{Structure of a material agent\label{fig:matAgent}}
\end{figure*}
\par\end{center}

As shown in Fig. \ref{fig:matAgent}, the material agent iteratively
carries out the following tasks:
\begin{enumerate}
\item check planned downstream consumption, update demand-supply situation
\item when production cannot fulfill all demands before required date, try
to find best solution by performing optimization from the supplier
material agent's perspective to allocate its available material to
each customer
\item event-driven: the agent receives a change request from its upstream
supplier agents
\item consolidate all change requests and try to find best solution by performing
optimization from the customer material agent's perspective to reconcile
change proposals from its suppliers
\item when all constraints are met, check for potentially excessive inventory
and reduce production accordingly
\end{enumerate}

\subsubsection{Supplier Material Agent's Optimization of Schedule \label{subsec:Supplier-Material-Agent's}}

First optimize from a supplier's perspective: a material agent notices
its own produced material quantity cannot fulfill all downstream customer
agents' orders before their demand date. It tries to match as much
demand as possible based on the given order priority and the given
strategy of order fulfillment: whether it is allowed to fulfill only
partial orders.

If partial order fulfillment is allowed, its objective is defined
as maximize (using $M$: number of orders; $N$: rescheduling horizon;
$d_{mn}$: demand of order m on day n; $s_{n}$: maximum available
supply on day n; $x_{mn}$: quantity allocated to order m on day n;
$W_{m}$: weight, or priority of order m): $obj=\sum\limits _{m=1}^{M}\sum\limits _{n=1}^{N}W_{m}*x_{mn}$
, with the demand constraints $\sum\limits _{n=1}^{N}x_{mn}<=\sum\limits _{n=1}^{N}d_{mn},\forall m\in[1,M]$
and supply constraints $\sum\limits _{m=1}^{M}x_{mn}<=\sum\limits _{i=1}^{n}s_{i},\forall n\in[1,N]${\small{}.}{\small\par}

The weight $W_{m}$ is order-priority related. The higher the priority,
the higher the weight. It can be linear or nonlinear to priority.
The weight used in the project increases linearly with priority.

If orders can only be delivered in full, or be canceled, the agent's
objective is defined as maximize $\sum\limits _{m=1}^{M}\sum\limits _{n=1}^{N}F_{m}*x_{mn}+\sum\limits _{m=1}^{M}\sum\limits _{n=1}^{N}L_{m}*f_{mn}$,
where $f_{mn}=x_{mn}*(x_{mn}-d_{mn})${\small{}} makes sure the adjusted
production quantity is either 0 or the original order quantity. Demand
and supply constraints are the same as in the previous problem. {\small{}$L_{m}$}
is the weight of planning adherence. It punishes date changes to the
original order and encourages that the existing schedule is kept as
much as possible{\small{}. $F_{m}$ is the} weight of order fulfillment.
It punishes partially fulfilled orders and encourages orders being
produced in full quantity.
\begin{center}
{\small{}}
\begin{figure*}
\begin{centering}
{\small{}\includegraphics[width=1\textwidth]{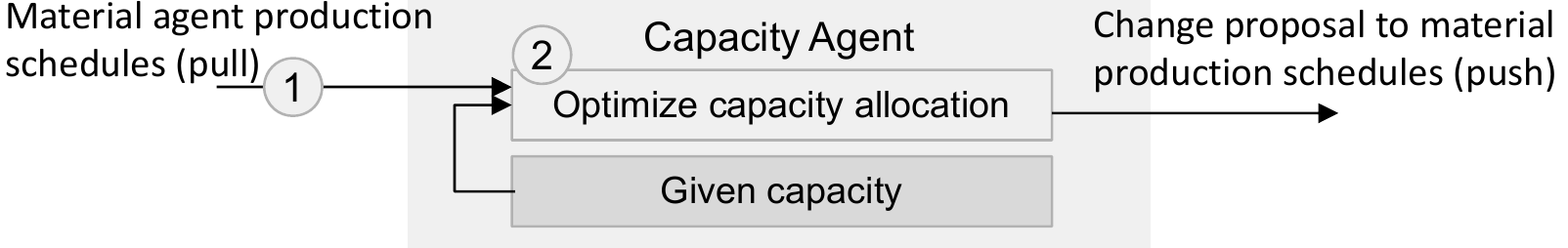}}{\small\par}
\par\end{centering}
{\small{}\caption{{\small{}Structure of a capacity agent\label{fig:capaAgent}}}
}{\small\par}
\end{figure*}
{\small\par}
\par\end{center}

\subsubsection{Customer Material Agent's Consolidation of Proposals \label{subsec:Customer-Material-Agent's}}

From the customer's perspective a material agent receives multiple
schedule change requests from its upstream supplier agents. This is
typical in real-world settings where, for example, a logistic disruption
result in multiple suppliers delaying delivery, although impact varies
based inventory level and severity of disruption. To avoid unnecessarily
expediting incoming materials (e.g. since production is already delayed
due to other parts from additional suppliers are missing), customer
agents should reschedule based on minimum availability.

The agent's objective is to minimize (using $M$: number of orders;
$N$: rescheduling horizon; $q_{mn}$: requested quantity reduction
or reschedule of order m on day n; $r_{mn}$: reduced or rescheduled
quantity of order m on day n; $W_{mn}$: weight of order m on day
n): $obj=\sum\limits _{n=1}^{N}W_{mn}r_{mn}$ with the constraints
$\sum\limits _{i=1}^{n}r_{i}>=\sum\limits _{i=1}^{n}q_{mi},\forall m\in[1,M],n\in[1,N]$.
The constraints ensure that at any time step the customer agent's
accumulated reduced production quantity is greater than or equal to
the proposed reduction quantity from any one of the supplier agents.
{\small{}$W_{mn}$ }attenuates later orders and encourages solutions
which schedule production as early as possible.

\subsection{Capacity Agent}

As shown in Fig. \ref{fig:capaAgent}, the capacity agent iteratively
carries out the following tasks:
\begin{enumerate}
\item Check planned downstream consumption, update demand-supply situation
\item If capacity cannot fulfill demand, optimize and propose changes to
its downstream agents
\end{enumerate}

\subsubsection{Capacity Agent's Consolidation of Proposals \label{subsec:Capacity-Agent's-Consolidation}}

For a capacity agent, the optimization problem has the objective of
maximizing (using $M$: number of orders; $N$: rescheduling horizon;
$d_{mn}$: demand of order m on day n; $s_{n}$: maximum capacity
on day n; $x_{mn}$: quantity allocated to order m on day n): $obj=\sum\limits _{m=1}^{M}\sum\limits _{n=1}^{N}x_{mn},$
with the maximum demand constraints $\sum\limits _{n=1}^{N}x_{mn}<=\sum\limits _{n=1}^{N}d_{mn},\forall m\in[1,M]$
and maximum capacity constraints $\sum\limits _{m=1}^{M}x_{mn}<=s_{n},\forall n\in[1,N]$.

\section{Implementation}

The simulation system has been implemented using the Jadex Active
Components framework \cite{Braubach+12Developing}. Jadex allows for
implementing distributed agent systems using an overlay network of
agent platforms, that automatically discover each other depending
on virtual network settings \cite{Braubach+13Security}. In case of
scenarios with deep and complex BOMs, this allows for horizontal scaling
by adding platforms according to the needed number of agents. An application
built with Jadex can be executed in a local and distributed setup
without requiring code modifications as communication between agents
is done via remote service interfaces. The agent implementation in
Jadex can be done following different agent architectures, i.e. BDI
(belief-desire-intention) \cite{Rao+95} or simple task-based agents.
In any case agents are realized using the unmodified Java programming
language so that well-known IDEs like Eclipse can further be used.

The implementation uses a main program that reads Huawei BOM data
and creates material and capacity agents accordingly. The material
agents will be linked to each other so that each agent knows their
parent and child agents. Similarly, material and capacity agents are
connected when a production relationship between them exists. Once
the setup is completed, the system offers a user interface to initiate
different disruption events. To realize the interactions between the
agents a service oriented approach has been used, i.e. each agent
offers a service that allows for interacting with other agents. In
this case, for each agent type one service - called IMaterialService
and ICapabilityService - was devised. 

\begin{figure}[h]
\begin{lstlisting}
@Service 
public interface IMaterialService {
  public IFuture<SupplyData> getSupply();
  public IFuture<DemandData> getDemand();
  public IFuture<Void> addSupply(SupplyData d);
  public IFuture<Void> removeSupply(SupplyData d);
  public IFuture<Void> addDemand(DemandData d);
  public IFuture<Void> removeDemand(DemandData d);
  public IFuture<Void> handleChangeProposal(ChangeProposal p);
  public IFuture<Schedule> optimizeLocalSchedule();
  public IFuture<SupplyData> getSupplyGap();
  public IFuture<Void> applyDisruption(DisruptionEvent e);
}
\end{lstlisting}
\caption{Jadex material service interface}
\label{fig:matservice} 
\end{figure}

In Fig. \ref{fig:matservice} the IMaterialService is depicted. It
can be seen that the service interface starts with a @Service annotation,
which is a marker interface for Jadex that the definition is a remotely
callable service. It can also be observed that all methods are declared
asynchronously using an IFuture return value. This means that the
caller and callee agent are decoupled and the caller will not block
until the return value arrives. The real return types are visible
as generic types in the future declarations. The interface allows
for fetching the current state of the local tasks using getSupply()
and getDemand() methods. Furthermore, the state can be changed by
adding or removing supply and demand using the following methods.
In case another agents intends to influence the internal planning
state of the agent, it can do so by calling the handleChangeProposal()
method. 

Even though in the current prototype, material agents optimize their
local schedule only when triggered internally, the service also has
a optimizeLocalSchedule() method, which allows for other agents or
the user interface to initiate a local replanning. Finally, the supply
gap can be fetched and disruption events can be dispatched to the
agent. Receiving such an event, the agent will start handling this
kind of disruption by internally optimizing and escalating to other
agents as far as necessary.

In Fig. \ref{fig:capservice} the ICapabilityService is depicted.
It allows for adding and removing production task using the corresponding
add/removeProductionTask methods. In addition the local production
schedule at the production site can be obtained using getSchedule().
Similar as in material service, also the capability service offers
a method to initiate the local optimization of the capability via
optimizeLocalSchedule(). Finally, also the capability service accepts
disruption events using applyDisruptionEvent() so that disruption
scenarios at a concrete production site can be imitated.

\begin{figure}[h]
\begin{lstlisting}
@Service 
public interface ICapacityService { 	
  public IFuture<Void> addProductionTask(ProductionTask t); 
  public IFuture<Void> removeProductionTask(ProductionTask t); 
  public IFuture<Schedule> getSchedule(); 
  public IFuture<Void> optimizeLocalSchedule(); 
  public IFuture<Void> applyDisruption(DisruptionEvent e); }
\end{lstlisting}
\caption{Jadex capability service interface}
\label{fig:capservice} 
\end{figure}

\begin{table*}
\caption{System performance with different disruption durations\label{tbl:disruptionduration}}
\begin{tabular}{>{\centering}m{0.1\columnwidth}>{\centering}m{0.1\columnwidth}>{\centering}m{0.1\columnwidth}>{\centering}m{0.1\columnwidth}>{\centering}m{0.1\columnwidth}>{\centering}m{0.1\columnwidth}>{\centering}m{0.1\columnwidth}>{\centering}m{0.1\columnwidth}>{\centering}m{0.1\columnwidth}}
\hline 
{\small{}disruption type} &
{\small{}disruption duration} &
{\small{}average no. of iterations} &
{\small{}avg. no. of resched. material agents} &
{\small{}avg. no. of resched. capacity agents} &
{\small{}average number of resched. FGs} &
{\small{}avg. FG order fulfillment rate by orders} &
{\small{}avg. FG order fulfillment rate by volume} &
{\small{}maximum delay of FG orders}\tabularnewline
\hline 
\hline 
\multirow{5}{0.1\columnwidth}{{\small{}line stoppage}} &
{\small{}1} &
{\small{}5} &
{\small{}14} &
{\small{}13} &
{\small{}5} &
{\small{}99.41\%} &
{\small{}99.99\%} &
{\small{}1}\tabularnewline
 & {\small{}3} &
{\small{}5} &
{\small{}14} &
{\small{}13} &
{\small{}6} &
{\small{}99.41\%} &
{\small{}99.99\%} &
{\small{}3}\tabularnewline
 & {\small{}5} &
{\small{}7} &
{\small{}14} &
{\small{}13} &
{\small{}6} &
{\small{}99.41\%} &
{\small{}99.98\%} &
{\small{}5}\tabularnewline
 & {\small{}7} &
{\small{}7} &
{\small{}15} &
{\small{}15} &
{\small{}7} &
{\small{}99.41\%} &
{\small{}99.98\%} &
{\small{}8}\tabularnewline
 & {\small{}9} &
{\small{}7} &
{\small{}15} &
{\small{}15} &
{\small{}8} &
{\small{}98.83\%} &
{\small{}99.97\%} &
{\small{}9}\tabularnewline
\hline 
\multirow{5}{0.1\columnwidth}{{\small{}raw material stoppage}} &
{\small{}1} &
{\small{}1} &
{\small{}2} &
{\small{}3} &
{\small{}2} &
{\small{}99.42\%} &
{\small{}99.32\%} &
{\small{}1}\tabularnewline
 & {\small{}3} &
{\small{}2} &
{\small{}6} &
{\small{}7} &
{\small{}4} &
{\small{}98.55\%} &
{\small{}97.99\%} &
{\small{}3}\tabularnewline
 & {\small{}5} &
{\small{}2} &
{\small{}8} &
{\small{}8} &
{\small{}5} &
{\small{}97.38\%} &
{\small{}94.75\%} &
{\small{}5}\tabularnewline
 & {\small{}7} &
{\small{}3} &
{\small{}9} &
{\small{}9} &
{\small{}7} &
{\small{}96.51\%} &
{\small{}92.93\%} &
{\small{}7}\tabularnewline
 & {\small{}9} &
{\small{}3} &
{\small{}9} &
{\small{}9} &
{\small{}7} &
{\small{}96.51\%} &
{\small{}92.93\%} &
{\small{}9}\tabularnewline
\hline 
\multirow{5}{0.1\columnwidth}{{\small{}semi-finished goods quarantine}} &
{\small{}1} &
{\small{}3} &
{\small{}10} &
{\small{}7} &
{\small{}5} &
{\small{}99.41\%} &
{\small{}99.99\%} &
{\small{}1}\tabularnewline
 & {\small{}3} &
{\small{}4} &
{\small{}10} &
{\small{}8} &
{\small{}6} &
{\small{}99.12\%} &
{\small{}99.99\%} &
{\small{}3}\tabularnewline
 & {\small{}5} &
{\small{}4} &
{\small{}10} &
{\small{}7} &
{\small{}6} &
{\small{}99.12\%} &
{\small{}99.98\%} &
{\small{}5}\tabularnewline
 & {\small{}7} &
{\small{}4} &
{\small{}10} &
{\small{}7} &
{\small{}5} &
{\small{}99.41\%} &
{\small{}99.98\%} &
{\small{}7}\tabularnewline
 & {\small{}9} &
{\small{}4} &
{\small{}10} &
{\small{}8} &
{\small{}6} &
{\small{}99.41\%} &
{\small{}99.98\%} &
{\small{}9}\tabularnewline
\hline 
\end{tabular}
\end{table*}

This service-oriented architecture is meant to be flexible and serves
as foundation for the realization of different scheduling approaches.
The main workflow of the material agent described in Section \ref{subsec:Material-Agent}
is realized by the services as follows. As first step the demand situation
is updated by asking the parent material agents for the current demand
using the getDemand() method. The individual results will be integrated
and when demand exceeds supply, the distribution of parts to customers
is optimized locally. The new supply situation is communicated to
the consumers using handleChangeRequest() method. A material agent
that receives such a request waits for requests of all its parents
and consolidates the requests internally. Afterwards it sends its
possibly adjusted proposals back to the suppliers. A capability agent
uses the getSupply() and getDemand() methods of all participating
material agents in its schedule to update its current production situation.
When demand is not satisfied by the available supplies, the agent
optimizes locally and also uses the handleChangeRequest() method to
propose adapted productions.

\section{Simulation Results}

 The simulation considers five different BOM structures representing
five product groups with varying complexity (BOM hierarchies ranging
from 2 to 7 levels). The simulation data is based on real supply chain
data from Huawei. This simulation setting consists of a BOM model
resulting in 39 material agents and data on production facilities
for 18 capacity agents. p In the traditional model of manual rescheduling,
the benchmark for a successful recovery is that the compensation finishes
within a timeframe of up to 14 days from the day of disruption. In
addition, after the 14 days the new schedule must recover 100\% of
the planned production volume. Since these values are used as baseline
for evaluating the success of the manual reschedules are disruptions,
it can also be applied to the agent-based rescheduling approach. However,
the manual process may take anywhere between 1 to 7 days to complete
as planners need to wait overnight for manual adjustments to be reflected
in the new schedule after which they can decide on further adjustments
to mitigate ripple effects of the implemented changes.

Before the simulation begins, the production schedule is initialized
as a feasible but not optimal schedule with no disruptions. Three
disruption types are simulated:
\begin{itemize}
\item In a raw material shortage scenario, a raw material agent's transit
plan is delayed.
\item In a semi-finished goods quarantine scenario, a material agent in
the middle of the BOM tree has delayed availability.
\item In a line stoppage scenario, a capacity agent's available capacity
is reduced to zero for the duration of the disruption.
\end{itemize}
Each scenario is tested with three different inventory days-on-hand
levels - 14.8, 6.7 and 3.8 days. In each situation, disruption durations
ranging from 1 to 9 days are tested. One agent in each of the five
BOMs is directly impacted. Rescheduling is permitted within 14 days
of the first disruption after which unfulfilled orders will be canceled.
Two types of performance measurement indicators are chosen: the number
of iterations needed for the system to stabilize as well as the number
of affected material/capacity/finished goods agents are the speed
indicators with lower number of iterations and affected nodes meaning
the system can generate solutions faster. On the other hand, finished
goods order fulfillment rates in terms of number of orders and in
terms of total volume and maximum delay are indicators for the quality
of the solution. The simulation executed on a 2.7GHz/16GB Intel i7
2core/4thread machine with each iteration taking approximately 1 minute.

{\small{}}{\small\par}

{\small{}}
\begin{table*}
\center{\small{}\caption{System performance with different inventory levels\label{tbl:doh-2}}
}%
\begin{tabular}{>{\centering}p{0.115\columnwidth}>{\centering}p{0.115\columnwidth}>{\centering}p{0.115\columnwidth}>{\centering}p{0.115\columnwidth}>{\centering}p{0.115\columnwidth}>{\centering}p{0.115\columnwidth}>{\centering}p{0.115\columnwidth}>{\centering}p{0.115\columnwidth}}
\hline 
{\footnotesize{}disruption type} &
{\footnotesize{}avg. inventory days-on hand} &
{\footnotesize{}avg. no. of iterations} &
{\footnotesize{}avg. no. of rescheduled material agents} &
{\footnotesize{}avg. no. of rescheduled capacity agents} &
{\footnotesize{}avg. no. of rescheduled FGs} &
{\footnotesize{}avg. FG order fulfillment rate by orders} &
{\footnotesize{}avg. FG order fulfillment rate by volume}\tabularnewline
\hline 
\hline 
\multirow{3}{0.115\columnwidth}{{\footnotesize{}line stoppage}} &
{\small{}14.8} &
{\small{}6} &
{\small{}12} &
{\small{}14} &
{\small{}4} &
{\small{}100.00\%} &
{\small{}100.00\%}\tabularnewline
 & {\small{}6.7} &
{\small{}5} &
{\small{}14} &
{\small{}14} &
{\small{}6} &
{\small{}99.83\%} &
{\small{}100.00\%}\tabularnewline
 & {\small{}3.8} &
{\small{}6} &
{\small{}17} &
{\small{}14} &
{\small{}9} &
{\small{}98.05\%} &
{\small{}99.96\%}\tabularnewline
\hline 
\multirow{3}{0.115\columnwidth}{{\footnotesize{}raw material stoppage}} &
{\small{}14.8} &
{\small{}3} &
{\small{}5} &
{\small{}6} &
{\small{}4} &
{\small{}98.10\%} &
{\small{}95.83\%}\tabularnewline
 & {\small{}6.7} &
{\small{}1} &
{\small{}3} &
{\small{}4} &
{\small{}3} &
{\small{}97.93\%} &
{\small{}96.27\%}\tabularnewline
 & {\small{}3.8} &
{\small{}3} &
{\small{}12} &
{\small{}11} &
{\small{}8} &
{\small{}96.99\%} &
{\small{}94.65\%}\tabularnewline
\hline 
\multirow{3}{0.115\columnwidth}{{\footnotesize{}SFG quarantine}} &
{\small{}14.8} &
{\small{}3} &
{\small{}7} &
{\small{}7} &
{\small{}3} &
{\small{}100.00\%} &
{\small{}100.00\%}\tabularnewline
 & {\small{}6.7} &
{\small{}3} &
{\small{}10} &
{\small{}7} &
{\small{}5} &
{\small{}99.66\%} &
{\small{}100.00\%}\tabularnewline
 & {\small{}3.8} &
{\small{}5} &
{\small{}13} &
{\small{}8} &
{\small{}9} &
{\small{}98.23\%} &
{\small{}99.96\%}\tabularnewline
\hline 
\end{tabular}
\end{table*}
{\small\par}

As illustrated in Table \ref{tbl:disruptionduration}, in general,
both the speed indicators and the quality indicators deteriorate as
disruption duration increases. However, the system is able to recover
almost all finished goods orders in the line stoppage and semi-finished
goods quarantine scenarios. In all cases but one, maximum delay of
fully delivered finished goods orders does not exceed disruption duration.
In the raw material shortage scenario the system is performing slightly
worse but still manages to achieve more than 92\% coverage in total
volume and more than 96\% in total number of orders. The worst solution
is mostly due to impact beyond the 14-day rescheduling time window.
Additional improvements could be achieved using different solvers
and parameter settings in the models. The difference regarding affected
nodes across three disruption types reflects reality on the shop-floor:
in case of a line stoppage, utilization of production capacity is
severely impacted. As shown in Table \ref{tbl:disruptionduration},
regardless of disruption duration, a majority of the 18 capacity agents
are affected in the line stoppage scenario. Raw material shortage
tends to affect less material nodes compared to semi-finished goods
quarantine due to inventory build-up along the BOM: additional layers
of inventory buffer help dampening the impact from leaf nodes. Table
\ref{tbl:doh-2} depicts how different levels of inventory buffers
can affect speed and quality of the solution. In scenarios of raw
material shortage and line stoppage, both speed and quality indicators
worsen as inventory level drops. In the case of a raw material shortage,
results are mixed. An average inventory level of 6.7 days on-hand
is performing better than a higher inventory level. This may be due
to the initialization condition; however more investigation is needed
to fully understand the underlying reasons.

In general the system does not have the goal of providing a perfect
substitute schedule nor providing a replacement for the regular scheduling
system running in 24 hour cycles. Rather the system aims to provide
planners with suggestions that have minimal ripple effects on the
rest of the schedule, thereby reducing the number of days required
for replanning. In addition the system provides planners with more
instantaneous feedback on their own proposed changes, allowing for
higher confidence in the decisions made. The success of the system
in finding a solution could also be improved by allowing the system
additional resources. Currently the evaluation was performed by a
single system, but due the system consisting of distributed agents,
additional machine or clusters of machines could be employed to add
more computational resources to the system. Furthermore, additional
iterations may provide even better solutions, provided additional
time or computational resources can be allotted. A trade-off can be
considered here between the quality of the solution and the requirement
of speedy feedback for the planners.

\section{Conclusion}

In this paper a novel agent-based scheduling mechanism has been presented
in order to deal with supply chain disruptions. In contrast to existing
planners that produce optimal solutions but need considerable time,
here an approach is used that only adapts the existing schedule to
some degree depending on the severity of the disruptions. The approach
does not aim at an optimal solution but instead tries to fix occurring
problems via fast and efficient partial replanning. Conceptually,
the domain is modelled with material agents representing the parts
of different stages and capacity agents representing the production
sites. Material agents are created as hierarchy resembling BOM production
structures and capacity agents are instantiated according to real
production sites. The mechanism has been implemented as simulation
system using the Jadex agent platform.

Simulations have been performed based on real production schedule,
transit, capacity and parts hierarchy data of 57 nodes (39 material
parts and 18 production lines), with varying disruption severity and
inventory buffers. Speed indicators reveal the system being capable
of rescheduling with few iterations, limited change to initial schedule
and a low number of affected nodes. Quality indicators indicate that
rescheduling solutions are slightly worse but comparable to optimal
solutions of recovering all orders in full. These results imply that
such a simulation system can be used in complex supply chain setups,
offering quick decision support in case of a disruption as well as
supporting long-term supply chain design changes. Although not yet
fully tested with large-scale problems, the proposed agent-based system's
distributed nature indicates good scalability. Performance is directly
related to the complexity of the BOM hierarchy - a higher number of
layers or number of connections from any single node will increase
the number of iterations needed to reach stability. On the other hand,
if the rescheduling horizon is extended and the system is allowed
more flexibility to reschedule, quality of the solution is expected
to improve while the number of iterations is expected to increase.

Future work will investigate simulations of large-scale scenarios
and different kinds of scheduling disruptions as well as different
optimization algorithms such as learning algorithms based on historical
data, heuristics or rule-based engines within each agent. Additionally,
the simulation prototype will be extended in a current cooperation
project at Huawei.

\end{document}